\documentclass{article}

\usepackage{microtype}
\usepackage{graphicx}
\usepackage{subfigure}
\usepackage{booktabs} 

\usepackage{hyperref}

\usepackage[accepted]{icml2023}

\usepackage{amsmath}
\usepackage{amssymb}
\usepackage{mathtools}
\usepackage{amsthm}

\usepackage[capitalize,noabbrev]{cleveref}

\theoremstyle{plain}

\theoremstyle{definition}

\theoremstyle{remark}

\usepackage[textsize=tiny]{todonotes}

\usepackage{xspace}
\usepackage{multirow}

\makeatletter
\DeclareRobustCommand\onedot{\futurelet\@let@token\@onedot}
\def\@onedot{\ifx\@let@token.\else.\null\fi\xspace}

\def\eg{\emph{e.g}\onedot}

\def\etc{\emph{etc}\onedot} \def\vs{\emph{vs}\onedot}

\makeatother

\icmltitlerunning{Img2Vec: A Teacher of High Token-Diversity Helps Masked AutoEncoders}

\begin{document}

\twocolumn[
\icmltitle{Img2Vec: A Teacher of High Token-Diversity Helps Masked AutoEncoders}



\icmlsetsymbol{equal}{*}

\begin{icmlauthorlist}
\icmlauthor{Heng Pan}{equal,tx}
\icmlauthor{Chenyang Liu}{equal,tx}
\icmlauthor{Wenxiao Wang}{zju}
\icmlauthor{Li Yuan}{pku}
\icmlauthor{Hongfa Wang}{tx}
\icmlauthor{Zhifeng Li}{tx}
\icmlauthor{Wei Liu}{tx}

\end{icmlauthorlist}

\icmlaffiliation{pku}{School of Electrical and Computer Engineering, Peking University, Shenzhen, China}
\icmlaffiliation{tx}{Data Platform, Tencent, Shenzhen, China}
\icmlaffiliation{zju}{School of Software Technology, Zhejiang University, Hangzhou, China}

\icmlcorrespondingauthor{Zhifeng Li}{michaelzfli@tencent.com}
\icmlcorrespondingauthor{Wei Liu}{wl2223@columbia.edu}


\vskip 0.3in
]



\printAffiliationsAndNotice{\icmlEqualContribution} 

\begin{abstract}

We present a pipeline of Image to Vector (Img2Vec) for masked image modeling (MIM) with deep features. To study which type of deep features is appropriate for MIM as a learning target, we propose a simple MIM framework with serials of well-trained self-supervised models to convert an Image to a feature Vector as the learning target of MIM, where the feature extractor is also known as a teacher model. Surprisingly, we empirically find that an MIM model benefits more from image features generated by some lighter models (\eg, ResNet-50, 26M) than from those by a cumbersome teacher like Transformer-based models (\eg, ViT-Large, 307M). To analyze this remarkable phenomenon, we devise a novel attribute, token diversity, to evaluate the characteristics of generated features from different models. Token diversity measures the feature dissimilarity among different tokens. Through extensive experiments and visualizations, we hypothesize that beyond the acknowledgment that a large model can improve MIM, a high token-diversity of a teacher model is also crucial. Based on the above discussion, Img2Vec adopts a teacher model with high token-diversity to generate image features. Img2Vec pre-trained on ImageNet unlabeled data with ViT-B yields 85.1\% top-1 accuracy on fine-tuning. Moreover, we scale up Img2Vec on larger models, ViT-L and ViT-H, and get $86.7\%$ and $87.5\%$ accuracy respectively. It also achieves state-of-the-art results on other downstream tasks, \eg, 51.8\% mAP on COCO and 50.7\% mIoU on ADE20K. Img2Vec is a simple yet effective framework tailored to deep feature MIM learning, accomplishing superb comprehensive performance on representative vision tasks.

\end{abstract}

\section{Introduction} \label{sec:intro}

Masked image modeling (MIM)\cite{bao2021beit, he2022masked, xie2022simmim} has become a new paradigm to pre-train vision transformer\cite{dosovitskiy2020image,bai2022improving} models. During the pre-training process, part of an input image is masked and a pre-trained model is then designed to reconstruct it by utilizing the unmasked part. MIM learning requires the model to understand the content of the input image. Thus, a certain pre-trained model which can correctly predict the masked part is believed to perform well on downstream tasks.

The critical point of MIM is the choice of reconstruction targets as it determines what we expect the model to learn. For example, MAE\cite{he2022masked}, \etc. reconstruct raw pixels or low-level (\eg, histograms of oriented gradients, HOG \cite{dalal2005histograms}) features of the masked part. While other methods (\eg, data2vec\cite{baevski2022data2vec}, MVP\cite{wei2022mvp} and iBOT\cite{zhou2021ibot}) resort to high-level semantic features for 
reconstruction targets. Compared with raw pixels or low-level features, high-level feature reconstruction is closer to the needs of downstream tasks, thus yielding superior performance.
The semantic features may be generated by another pre-trained model or the exponential moving average (EMA) counterpart of the pre-training one. 

\begin{figure}[t]
  \centering
  \includegraphics[width=0.98\linewidth]{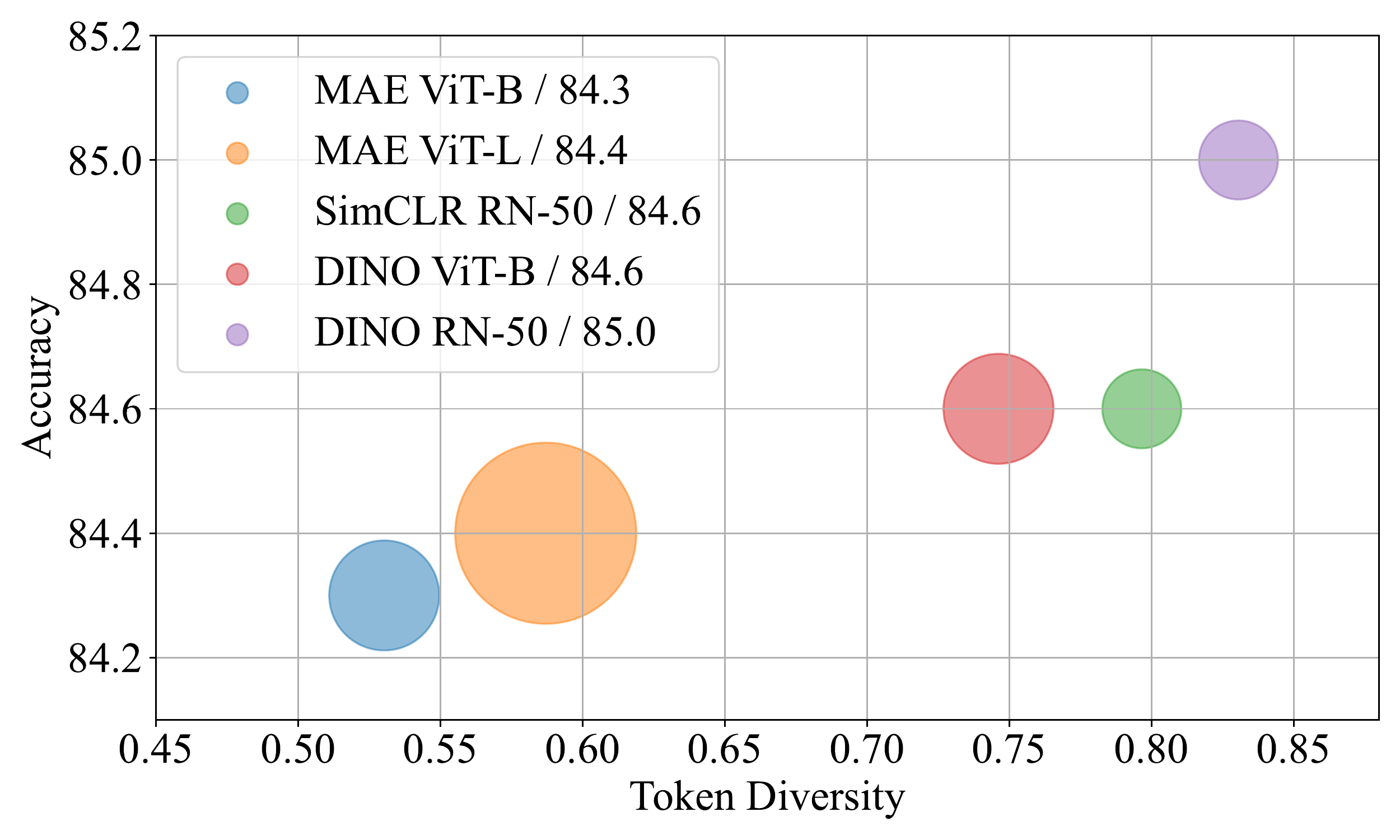}
  \vspace{-5pt}
  \caption{\textbf{MIM performance with various teachers.} We report the student's performance on ImageNet-1K with five pre-trained teachers. The area of a circle indicates the teacher's model size. A light model with high token-diversity, DINO ResNet-50, is more favorable for pre-training MIM.}
  \label{fig:core}
  \vspace{-10pt}
\end{figure}

Although these aforementioned algorithms have achieved great success, there hasn't yet existed a detailed study showing what high-level features are most effective for pre-training. To fill this gap, we first build a masked auto-encoder framework and carefully compare different kinds of reconstruction targets. Specifically, we take a masked image as the model's input and semantic features from other pre-trained models as reconstruction targets for pre-training. The pre-training model's (called student) performance is evaluated under the ImageNet-1K benchmark (\cref{fig:core}). We find that when taking a pre-trained model (called teacher) as the target provider, a better performing teacher does not necessarily lead to a better student, which is counter-intuitive and inconsistent with the commonsense from the knowledge distillation field.

Through qualitative and quantitative analyses, we attribute our observation to the token diversity of reconstruction targets. Similar to feature diversity\cite{zhang2022mask} that suggests the dispersion between samples, token diversity implies the discrimination between tokens. If the tokens of targets share high similarities, the model would be poor at distinguishing different semantics and be easy to converge, which is harmful to MIM. We argue that different from knowledge distillation, the output token-diversity of a teacher can be more critical than its model capacity to some extent for MIM.

Based on our discovery, we propose a simple yet effective pre-training framework named Img2Vec, which converts input \textbf{im}a\textbf{g}es \textbf{to} \textbf{vec}tors for masked part reconstruction. Similar to other self-distillation based MIM methods, Img2Vec also adopts a student-teacher learning style in pre-training. The student takes the visible patches of the masked image as input and extracts its features through an encoder. Then, a lightweight decoder follows to recover the semantic features of the masked part. The teacher takes the intact image as input and generates a sequence of tokens with the same length as the student. According to the token-diversity measurement, we choose the ResNet-50 pre-trained by DINO\cite{caron2021emerging} as our teacher. The inference cost of ResNet-50 during pre-training is friendly, thus making us scale our Img2Vec to larger models smoothly.
In addition, to promote the image-level visual understanding which is needed on some downstream tasks, we propose to introduce global semantics in our Img2Vec. We aggregate the student encoder's visible output to a global prediction, and let it be close to the global representation aggregated from the teacher.
To let different-level contents participate in masked part reconstruction, we introduce multi-block feature learning that aggregates features across the encoder's blocks to the decoder.
Finally, Img2Vec takes the encoder of the student for downstream tasks after pre-training. 
Unlike existing works, the teacher in Img2Vec is replaceable, regardless of architecture or parameter. It endows the teacher with great freedom of model choice, and a well-trained model also makes the pre-training process stable compared to EMA-based methods.

We conduct extensive experiments on downstream tasks such as image classification, object detection, semantic segmentation, and instance segmentation. The results show that Img2Vec outperforms existing state-of-the-art self-supervised pre-training methods on all these tasks. Specifically, with a vanilla ViT-Base model, our Img2Vec achieves $85.1\%$ top-1 accuracy on ImageNet-1K without extra data.


It is worth highlighting our contributions as follows:
\vspace{-10pt}
\begin{itemize}
    \item We provide a new perspective on the targets of MIM and analyze various deep features that served as learning targets. \vspace{-5pt}
    \item We provide a diversity-based principle to choose a good teacher for MIM. Based on this principle, a smaller model with high token-diversity can perform better than a much larger one. \vspace{-5pt}
    \item We propose a novel self-supervised pre-training framework for vision transformer named Img2Vec, which uses a small ResNet-50 as a teacher. Besides, we introduce multi-block feature learning and global semantic learning to enhance the Img2Vec. \vspace{-5pt}
    \item Extensive experiments are conducted on downstream tasks like image classification, object detection, and semantic segmentation to show the effectiveness of Img2Vec over the previous state-of-the-art methods. \vspace{-5pt}
\end{itemize}

\begin{figure*}
  \centering
  \includegraphics[width=0.96\textwidth]{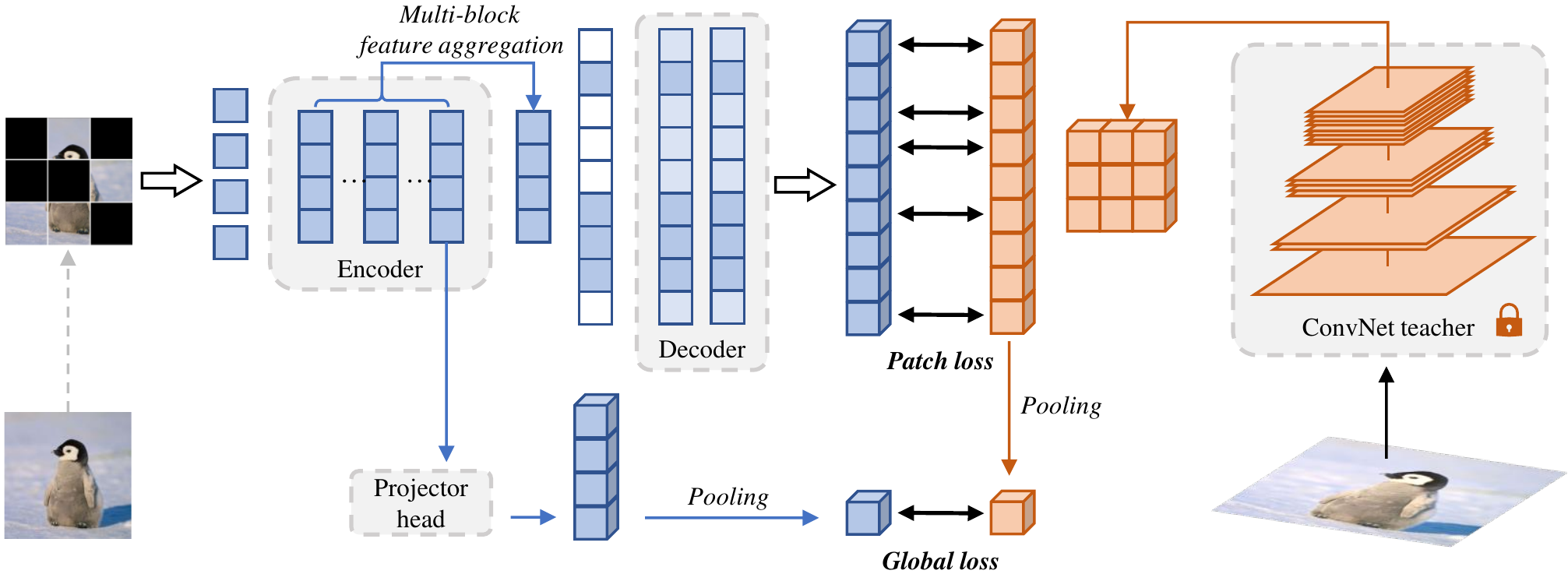}
  \vspace{-5pt}
  \caption{\textbf{The illustration of our Img2Vec pipeline.} It uses an asymmetric encoder-decoder architecture to predict the features of masked patches. The features are provided by a high token-diversity ConvNet teacher whose parameters are frozen. In addition, Img2Vec uses multi-block feature aggregation for efficient training and extracts a global feature of input through a 2-layer projector head to bring image-level supervision to the model.}
  \label{fig:Img2Vec}
  \vspace{-10pt}
\end{figure*}

\section{Related works}

\textbf{Self-supervised representation learning} uses abundant non-annotated data to prompt the model to learn a good representation by optimizing specific pretext tasks, which becomes an indispensable method of training large-scale models with limited labeled data. These methods can be broadly divided into two main categories according to the difference of pretext tasks: \textit{discriminative} and \textit{generative}. 
Discriminative approaches solve tasks like predicting rotation angle\cite{komodakis2018unsupervised}, finding the permutation of jigsaw puzzle\cite{noroozi2016unsupervised}, and masked object classification\cite{su2019vl}. These approaches are outdated and poor on downstream tasks.
Another part of discriminative approaches is summed up to contrastive learning, whose renowned methods are SimCLR\cite{chen2020simple}, MoCov3\cite{chen2021empirical}, BYOL\cite{grill2020bootstrap}, SimSiam\cite{chen2021exploring}, DINO\cite{caron2021emerging}, and so on. They aim to solve the instance discrimination task that encourages maximizing the similarity between two correlated views of one data sample and minimizing the similarity of the others. Contrastive-based methods are heavily dependent on data augmentation and batch size, which are difficult to scale up to large models.
Generative approaches aim to synthesize the input corrupted images or videos through generation tasks such as  colorization\cite{goyal2019scaling} and the prevalent masked image modeling\cite{bao2021beit}.

\textbf{Masked image modeling} has achieved tremendous attention and progress in the past years. It enables the model to learn an implicit visual representation by recovering the corrupted images.
BEiT\cite{bao2021beit} is a pioneering method that uses MIM as the pretext task for model pre-training. It was inspired by BERT\cite{devlin2018bert} in natural language processing, which aims to predict masked patch tokens discretized by a visual tokenizer.
MAE\cite{he2022masked} builds a representative MIM framework. It reconstructs the original pixels of high-proportion masked images using an asymmetric encoder-decoder architecture, where the encoder only deals with visible patches, and the lightweight decoder operates on both latent representation and masked tokens.
Since then, studies of MIM have mainly focused on two directions: \textit{masking strategies} and \textit{reconstruction targets}.
While \cite{he2022masked,xie2022simmim,zhou2021ibot} use a simple random strategy, AttMask\cite{kakogeorgiou2022hide} and SemMAE\cite{li2022semmae} improve the masking strategy through the guidances of attention maps and semantic parts, respectively.
For reconstruction targets, follow-up works point out that recovery of raw RGB values will waste the capability on high-frequency details. 
MaskFeat\cite{wei2022masked} proposes to regress features of the masked area and chooses the hand-crafted histograms of oriented gradients (HOG)\cite{dalal2005histograms} feature as default targets.
Data2vec\cite{baevski2022data2vec} and iBOT\cite{zhou2021ibot} use deep features extracted by an offline network whose weights are updated by an exponential moving average (EMA) algorithm.
MVP\cite{wei2022mvp} uses a CLIP\cite{radford2021learning} model for feature extraction. 
These deep representation-based methods can be treated as knowledge distillation from a certain target model.
In this paper, we will show the vital role that deep feature plays in MIM, and discuss how to choose proper deep features in the next section.

\section{Method} \label{sec:method}

\begin{figure*}
	\centering
    \includegraphics[width=0.92\linewidth]{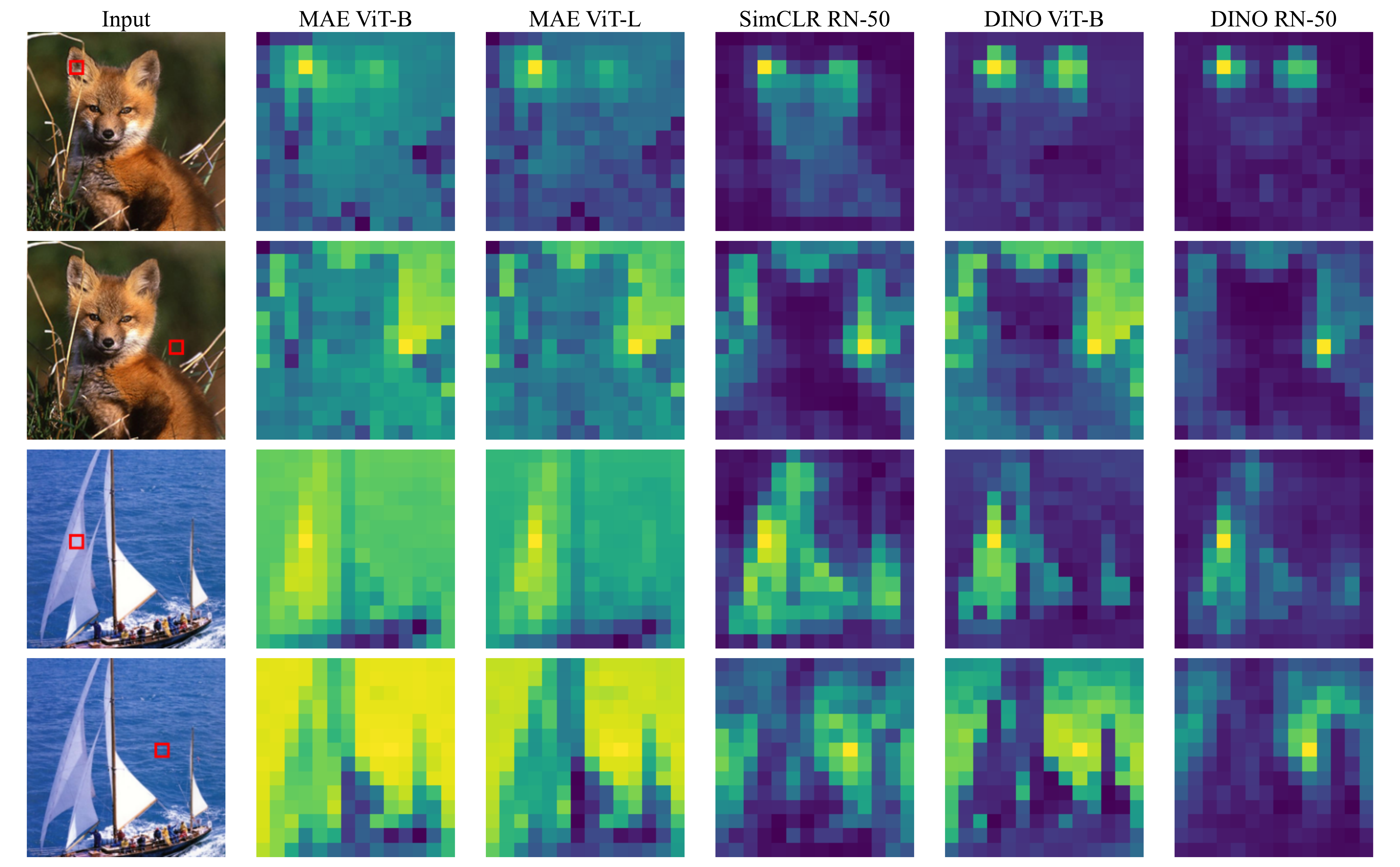}
    \vspace{-10pt}
	\caption{\textbf{Visualization of targets similarity from different teachers.} The area with a red box in the first column is the \textit{query} patch. The right 5 columns are similarity heat-maps of different teachers. Each grid indicates its similarity to the \textit{query}. The brighter the color, the greater the value, and vice versa.}
  \label{fig:teacher_token}
  \vspace{-15pt}
\end{figure*}

In this section, we first introduce a simple framework to fairly compare different features for MIM in Sec.~\ref{sec:dfbm}.
Based on the framework, we carefully compare reconstruction targets generated by different teachers and propose token diversity to evaluate their qualities in Sec.~\ref{sec:32}.
Then, we present our Img2Vec in Sec.~\ref{sec:image2vec}, which is equipped with a high token-diversity teacher for better MIM. This framework proposes two novel modules, multi-block feature learning, and global semantic learning, to improve the pre-training course. The whole framework is illustrated in \cref{fig:Img2Vec}.

\subsection{A unified framework for MIM} \label{sec:dfbm}

MIM has recently been an impressive self-supervised visual learning method and has drawn significant attention in the research community. Some works (\eg, MAE\cite{he2022masked} and MaskFeat\cite{wei2022masked}) use the pixel-level targets and reconstruct them from masked inputs. However, such a reconstruction target lacks of deep semantic information and performs worse than deep feature based methods.
Instead, more algorithms seek a deep feature based reconstruction target. For example, BEiT\cite{bao2021beit} trains a feature codebook first and optimizes a classification task on the masked tokens. While data2vec\cite{baevski2022data2vec} and BootMAE\cite{dong2022bootstrapped} adopt a momentum-updated model as a teacher to extract target features.  

Despite their success, the performance of the above methods (\eg, BEiT and data2vec) is affected by not only their chosen features but also their pre-training frameworks. It is difficult to compare the quality of different features fairly. To this end, we introduce a unified MIM framework.
The whole framework is similar to MAE\cite{he2022masked} except for the reconstruction target, which is extracted from a well-trained teacher model with the entire image as input. The teacher is frozen during MIM learning.

Since the teacher can be replaced by any model, we will study how to choose a better teacher for MIM in \cref{sec:32}.

\subsection{Token diversity for teacher evaluation} \label{sec:32}

\begin{table*}
  \centering
  \begin{tabular}{@{}clccccc@{}}
  \toprule
   & Teacher & Parameters & Linear probing & Fine-tuning & Token diversity & Student top-1 \\
  \midrule
  \multirow{2}{*}{MAE}    & ViT-Base/16   & 86M  & 68.0\% & 83.6\% & 0.530 & 84.3\% \\
                          & ViT-Large/16  & 307M & 75.8\% & 85.9\% & 0.587 & 84.4\% \\
  \midrule
  SimCLR                  & ResNet-50     & 26M  & 71.7\% & 76.3\% & 0.797 & 84.6\% \\
  \midrule
                          & ViT-Base/16   & 86M  & 78.2\% & 82.8\% & 0.746 & 84.6\% \\
  \multirow{-2}{*}{DINO}  & ResNet-50     & 26M  & 75.3\% & 77.4\%$^*$ & 0.831 & 85.0\% \\  
  \bottomrule
  \end{tabular}
  \caption{\textbf{Comparisons on teachers.} We present metrics on teacher models, including linear probing top-1  accuracy on ImageNet (Linear), fine-tuning accuracy (Finetune), token diversity, and pre-trained ViT-B accuracy (Student Top-1). DINO ResNet-50 has the highest token-diversity and the best student performance among all teachers, although its linear probing and fine-tuning accuracies are incomparable. $^*$The fine-tuning accuracy is our reproducing result, which is not reported in the original paper.}
  \label{tab:teacher}
  \vspace{-10pt}
\end{table*}

We conduct experiments with our pre-training framework to explore different reconstruction targets. For a fair comparison under the self-supervised setting, we assume that only the unlabeled ImageNet-1K\cite{russakovsky2015imagenet} is available, and no extra data is allowed during the pre-training process (thus teacher models pre-trained on CLIP\cite{radford2021learning} are not considered). We single out three well-known self-supervised models: MAE ViT-Base, MAE ViT-Large, and DINO ViT-Base.

MAE\cite{he2022masked} is the most representative pre-training method with MIM. It reconstructs the pixels of the masked patches and achieves more substantial results than contrastive learning based methods. DINO\cite{caron2021emerging} is a powerful contrastive learning method that aims to maximize the similarity between two views of a sample. It uses a student-teacher framework, where the teacher is updated by the EMA algorithm. It abandons negative pairs and achieves good performance.

The experimental settings are as follows:  We adopt ViT-B\cite{dosovitskiy2020image} as the student and pre-train it for $1600$ epochs. Following SimMIM\cite{xie2022simmim}, a random masking strategy with $60\%$ mask ratio and $32\times 32$ mask block size is adopted, and other settings are kept the same as MAE\cite{he2022masked}. When taking ViT-B and ViT-L as teacher models, we apply the same image size and patch size for both the student and teacher, which are $224^2$ and $16^2$, respectively. The output dimensions of ViT-B and ViT-L are $768$ and $1024$, respectively. The teachers' outputs are directly taken as the reconstruction targets.

The results are shown in \cref{tab:teacher}, and the last column presents the top-1 fine-tuning accuracy of the student they taught. As we can see, compared to the accuracy $84.3\%$ gotten by MAE ViT-B, using a larger MAE ViT-L as the teacher could slightly improve the performance by $0.1\%$. It can owe to the stronger feature extraction ability of ViT-L. In comparison, DINO ViT-B can bring a significant improvement compared to MAE ViT-B ($84.6\%$ \vs $84.3\%$) with the same teacher structure. This remarkable result indicates that there are more decisive factors for teacher selection than the model size and the teacher's performance. Hence the question arises: \textit{What are the most significant aspects that affect the teacher's quality?}

Since MIM is the task that uses the visible part to predict the masked part, the relations between different areas are worthy of being discussed. The role of a teacher is to provide reconstruction targets for the student, so we pay more attention to analyzing these targets. We visualize the relationship of target tokens by computing their pairwise cosine similarity, as shown in \cref{fig:teacher_token}. The area with a red box in the first column is the \textit{query} that calculates similarity with other tokens. The $2^{nd}$ and $5^{th}$ columns show the responses of MAE ViT-B and DINO ViT-B, respectively.

In \cref{fig:teacher_token}, we find that more tokens are similar to the \textit{query} in MAE ViT-B while being much less in DINO ViT-B. To give a quantitative analysis of this phenomenon, we present a metric named "token diversity". For a teacher feature sample with $K$ tokens $\{y_i\}_{i=1}^K$, we compute the pairwise cosine similarity between all tokens and use min-max normalization to map the value into $[0,1]$. Then we average these similarities for the sample value. Finally, we scan the whole ImageNet-1K training set with $N$ samples and get the diversity:
\begin{equation}
    diver = 1 - \frac{1}{N} \sum_{n=1}^N {sim}_n,
\end{equation}
where ${sim}_n=\frac{1}{K(K-1)} \sum_{i \ne j}^K \mathrm{norm} \{cosine(y_i, y_j)\}$ is the similarity of the $n^{th}$ sample. A larger $diver$ means that tokens distribute more diversely. 

The token diversities of MAE ViT-B, MAE ViT-L and DINO ViT-B are also reported in \cref{tab:teacher}. The comparison verified our assumption that DINO ViT-B with higher token-diversity teaches a better student.
In MIM pre-training, masked parts are reconstructed from visible parts.
If the masked targets and the visible targets are highly similar, predicting visible features themselves also satisfies the objective of MIM, which violates representation learning and compromises the performance of the student.
For example, if the patch of the fox's ear is masked, shown as the first row in \cref{fig:teacher_token}, the MAE-based teacher could tolerate the student to predict it as the fox's visible body feature, while the DINO-based teacher requires the student to predict more accurately. The token diversity suggests a discriminative ability to understand different visual areas.

Based on the token-diversity measurement, we question what model can get higher token-diversity. Considering that ViTs deal with all tokens in a global attention mechanism, their output tokens tend to be similar to each other. ConvNets are deemed to have the inductive bias of locality, which extract features with neighbors. Thus, ConvNets may be good alternatives for generating MIM targets. We choose the representative ConvNet, \eg ResNet-50 pre-trained by SimCLR and DINO, to calculate their token diversities in \cref{tab:teacher}, and visualize similarity heat-maps in \cref{fig:teacher_token}. The results show that the ResNet-50 gets higher token-diversity. It is worth noting that the more minor teacher, DINO ResNet-50, teaches a student who achieves $85.0\%$ top-1 accuracy on ImageNet-1k.




Through the above qualitative and quantitative analyses, we hypothesize that the token diversity is a valuable and essential perspective to measure the learning targets of MIM.


\subsection{Img2Vec} \label{sec:image2vec}

We introduce the Img2Vec framework for MIM tasks. According to the above discussion, we adopt the high token-diversity model, DINO ResNet-50, as the teacher. Furthermore, Img2Vec contains two novel modules, multi-block feature learning and global semantic learning. The overall framework is plotted in \cref{fig:Img2Vec}.

Our framework is based on MAE\cite{he2017mask} due to its simplicity and efficiency, but we replace the learning target from RGB values to deep features.
Concretely, the student encoder gathers all visible patches in a masked image, embeds them by a linear projection with position embeddings, and uses transformer blocks in tandem to extract latent tokens. Then, these visible tokens are merged with masked tokens, and are fed into a light decoder to predict the deep features of masked patches. The decoder also adopts transformer blocks, while its depth and width are smaller. In practice, we set the depth to 2 and 4 for ViT-B and ViT-L, respectively, and set its width to 512 for both. Finally, the decoder uses a linear layer to project the tokens to the target's dimension space (\eg, 2048 for the ResNet-50 teacher).

After obtaining the predictions and targets, we adopt the patch loss to optimize the model, which is similar to MAE\cite{he2022masked}. Smooth $L_1$ loss\cite{girshick2015fast} is used and only masked tokens are considered,
\begin{equation}
    \mathcal{L}_{\text{patch}} = \frac{1}{\Omega(\mathcal{M})} \sum_{i \in \mathcal{M}} \text{smooth}_{L_1} (y_i - z_i),
\end{equation}
where $y$ is the targets extracted by the teacher, $z$ is the predictions of the student, $\mathcal{M}$ is the set of masked patches indices, and $\Omega(\cdot)$ means the set cardinality. We set $\beta = 2$ in the smooth $L_1$ loss.

\textbf{ConvNet teacher.} As discussed in sec.\ref{sec:32}, DINO ResNet-50 has a high token-diversity and helps MIM tasks. Thus we adopt it in our Img2Vec. Different from the transformer, ResNet outputs a down-sampled feature map. The down-sample rate of ResNet-50 is $32$, while the encoder patch size is $16$. For example, when taking a $224^2$ image as the model input, the shape of the ResNet-50 final feature map is $2048\times7\times7$, while the encoder has $14\times14$ tokens. In order to keep the same feature shapes across the student and teacher, we do a $2\times$ up-sampling on teacher inputs with bilinear interpolation.
The final feature map of ResNet is adopted as the pre-training target.

\textbf{Multi-block feature learning.} In pre-training, an encoder-decoder structure is adopted. A typical practice only passes encoder's last block feature to the decoder, which neglects lower-level semantics. To make use of these semantics effectively, we propose multi-block feature learning. The representations of all encoder blocks are averaged as the outputs,
$h_{i} = \sum_{l=1}^L h_{i}^{l}$,
where $i$ denotes the patch index, and $l$ is the layer index.

\textbf{Global semantic learning.} MIM is the task that makes model learning through predicting the masked areas under patch-level supervision. This strategy may divert the model from understanding the visual sense in a whole image perspective, thus hurting the performance. We introduce global semantic learning to Img2Vec through an additional global loss. We calculate the smooth $L_1$ loss between the global embeddings of the teacher and student. 
The teacher's global embedding is the mean of $K$ tokens $\{y_i\}_{i=1}^K$. 
For the student, we first use a simple MLP, including two linear layers with one ReLU\cite{nair2010rectified} layer being a projector head $p$, to project $V$ visible tokens $\{h_i\}_{i=1}^V$ of the encoder's last layer to the teacher's dimension space. Then the encoder's global embedding is obtained by averaging these tokens, and the global loss is thus as:
\begin{equation}
\mathcal{L}_\text{global} = \text{smooth}_{L_1}(\frac{1}{\Omega(\mathcal{V})} \sum_{i \in \mathcal{V}}p(h_i) - \frac{1}{K}\sum_{i=1}^{K}y_i).
\end{equation}
$\mathcal{V}$ is the set of visible patches indices. We note that the [CLS] token does not participate in the computation for global embeddings.
The design of global semantics is simple and intuitive, and we get the global embedding in a similar manner as ViT does in the classification task, which takes an average pooling over the backbone's output as an image-level representation for the subsequent classifier.

Global semantics implicitly adopt two masked views of one image as a positive pair because of their same optimal objective. It desires a small distance between positive pairs, thus pushes the features from the same instance cluster closer.

Overall, the total loss of our Img2Vec is:
\begin{equation}
    \mathcal{L} = \mathcal{L}_\text{patch} + \lambda \mathcal{L}_\text{global}.
\end{equation}
$\lambda$ is a hyper-parameter to balance the global semantics.

\section{Experiments} \label{sec:exper}

\begin{table*}[t]
\centering
\begin{tabular}{@{}l|c|ccc|c|cc|c@{}}
\toprule
Method & Pre-train data & ViT-B & ViT-L & ViT-H & Linear   & AP$^\text{bbox}$ & AP$^\text{mask}$ & mIoU \\ 
\midrule
supervised                         & IN1K+labels & 82.3 & 82.6 & -    & -               & 47.9 & 42.9 & 47.4 \\
MoCo v3\cite{chen2021empirical}    & IN1K        & 83.2 & 84.1 & -    & 76.7$^\dagger$  & 47.9 & 42.7 & 47.3 \\
DINO\cite{caron2021emerging}       & IN1K        & 82.8 & -    & -    & 78.2$^\dagger$  & -    & -    & 47.2 \\
BEiT\cite{bao2021beit}             & IN1K+DALLE  & 83.2 & 85.2 & -    & 56.7            & 49.8 & 44.4 & 47.1 \\
MAE\cite{he2022masked}             & IN1K        & 83.6 & 85.9 & 86.9 & 68.0            & 50.3 & 44.9 & 48.1 \\
CAE\cite{chen2022context}          & IN1K        & 83.9 & 86.3 & -    & 70.4            & 50.0 & 44.0 & 50.2 \\
iBOT\cite{zhou2021ibot}            & IN1K        & 84.0 & 85.2 & -    & 79.5$^\dagger$  & -    & -    & 50.0 \\
SimMIM\cite{xie2022simmim}         & IN1K        & 83.8 & -    & -    & 56.7            & -    & -    & -    \\
MaskFeat\cite{wei2022masked}       & IN1K        & 84.0 & 85.7 & -    & -               & -    & -    & -    \\
data2vec\cite{baevski2022data2vec} & IN1K        & 84.2 & 86.6 & -    & -               & -    & -    & -    \\
BootMAE\cite{dong2022bootstrapped} & IN1K        & 84.2 & 85.9 & -    & 66.1            & 48.5 & 43.4 & 49.1 \\
SemMAE\cite{li2022semmae}          & IN1K        & 84.5 &   -  & -    & 68.7            & -    & -    & 46.3 \\
PeCo\cite{dong2023peco}            & IN1K        & 84.5 & 86.5 & 87.5 & -               & 47.8 & 42.6 & 48.5 \\
CMAE\cite{huang2022contrastive}    & IN1K        & 84.7 & 86.6 & -    & 73.9$^\dagger$  & -    & -    & 50.1 \\
\midrule
\textbf{Img2Vec } & \textbf{IN1K} & \textbf{85.1} & \textbf{86.7} & \textbf{87.5} & \textbf{75.8} & \textbf{51.8} & \textbf{46.0} & \textbf{50.7} \\ \bottomrule               
\end{tabular}
\vspace{-5pt}
\caption{\textbf{Comparisons with previous methods on various downstream tasks.} We report top-1 fine-tuning accuracies on ImageNet-1K at columns ``ViT-B", ``ViT-L" and ``ViT-H". For a fair comparison, all the above methods are only trained on the ImageNet training set. Column ``Linear" shows linear probing results on ImageNet-1K. Columns ``AP$^\text{bbox}$" and ``AP$^\text{mask}$" report the object detection and instance segmentation performances on COCO using Mask R-CNN as the detector. Last ``mIoU" column reports semantic segmentation results on ADE20k. UperNet is adopted as the segmentation framework. Note that except the ``ViT-L" and ``ViT-H" columns, all the results are reported on the ViT-B backbone; $^\dagger$ denotes the methods that used contrastive learning.}
\label{tab:my-table}
\vspace{-10pt}
\end{table*}

\subsection{Implementation details}

\textbf{Pre-training.} The pre-training setting follows MAE\cite{he2022masked}. We use a random mask strategy with a 60\% mask ratio and the mask size is $32\times 32$, which is adopted in SimMIM\cite{xie2022simmim}. Besides, we use AdamW\cite{loshchilov2018decoupled} as an optimizer, and momentum is set to $\beta_1=0.9, \beta_2=0.95$, and the weight decay is $0.05$. We adopt the linear scaling rule\cite{goyal2017accurate}, $lr=lr_{base}*batch\_size / 256$, and the base learning rate is set to $1.5\mathrm{e}{-4}$ with batch size $4096$. The entire training keeps $1600$ epochs. The cosine learning rate schedule\cite{loshchilov2016sgdr} with a $40$ epochs warm-up is adopted. Instead of using an 8-layer decoder in MAE\cite{he2022masked}, we use a lighter 2-layer decoder for ViT-Base and a 4-layer decoder for ViT-Large, respectively.


\subsection{Main results on ImageNet-1K}

Following previous works\cite{he2022masked, xie2022simmim, wei2022masked, bao2021beit, huang2022contrastive}, we use ImageNet-1K\cite{russakovsky2015imagenet} as the pre-training and fine-tuning dataset. ImageNet-1K contains 1.3M images with 1000 categories and is split into training and validation sets. We only use the training set during pre-training. After that, we fine-tune our model on the training set and evaluate its classification accuracy on the validation set. The fine-tuning settings also follow MAE\cite{he2022masked}. Since our pre-trained models generalize better than the previous methods, we decrease the base learning rate from $5\mathrm{e}{-4}$ to $1.25\mathrm{e}{-4}$ during fine-tuning. Besides, the common practices are adopted, \eg, mixup\cite{zhang2017mixup}, drop path\cite{huang2016deep}, layer decay\cite{clark2020electra}, and cutmix\cite{yun2019cutmix}.

In \cref{tab:my-table}, we compare Img2Vec against previous works on the ImageNet classification task. Img2Vec achieves $85.1\%$ top-1 accuracy with ViT-B, which surpasses MAE\cite{he2022masked} by $1.5\%$ absolutely. Compared with other deep feature-based methods (\eg, data2vec\cite{baevski2022data2vec} and BootMAE\cite{dong2022bootstrapped}), Img2Vec also outperforms by a remarkable gain. We also scale up our method on larger models, ViT-L, and get $86.7\%$ accuracy, which also beats MAE by $0.8\%$ absolutely. For a fair comparison, all the methods are pre-trained on the ImageNet training set only. Methods (\eg, MVP\cite{wei2022mvp}) using the CLIP\cite{radford2021learning} model are not mentioned.

\textbf{Linear probing.} We adopt the same setting of MAE\cite{he2022masked} to do linear probing. The results are shown in \cref{tab:my-table}. Img2Vec achieves $75.8\%$ top-1 accuracy. MoCoV3\cite{he2020momentum} and DINO\cite{caron2021emerging} use contrastive learning, which helps the features from different views cluster and thus encourages linear probing. CMAE\cite{huang2022contrastive} and iBOT\cite{zhou2021ibot} also employ contrastive learning. Without training on different views, our Img2Vec is only slightly inferior to contrastive methods but outperforms the other MIM-based methods.

\vspace{-5pt}
\subsection{Transfer learning experiments}

\textbf{Object detection and instance segmentation on COCO.} To verify the transferability of Img2Vec, we fine-tune the pre-trained model on COCO\cite{lin2014microsoft} for object detection and instance segmentation. Following MAE\cite{he2022masked}, we employ Mask RCNN\cite{he2017mask} with FPN\cite{lin2017feature} as the detector. The learning rate is decreased to $4\mathrm{e}{-5}$ with a cosine annealing schedule. The model is fine-tuned with 100 epochs.

The comparison results are shown in \cref{tab:my-table}. In both object detection and instance segmentation, Img2Vec achieves state-of-the-art performance under the same framework. It is noted that CMAE\cite{huang2022contrastive} conducts detection experiments with different detection heads, and that the results in the fair settings are not reported. MAE\cite{he2022masked} with this head achieves $51.7\%$ bbox mAP and $45.9\%$ mask mAP. iBOT adopts Cascade Mask R-CNN\cite{cai2019cascade}, which is also unfair.

\textbf{Semantic segmentation on ADE20K.} We also fine-tune our models on the ADE20K\cite{zhou2019semantic} dataset for semantic segmentation. Similar to previous methods, UperNet is adopted as the segmentation head. Same as the settings of MAE\cite{he2022masked}, we turn on relative position bias\cite{raffel2020exploring} during transfer fine-tuning. The learning rate is set to $2.5\mathrm{e}{-4}$ with $0.05$ weight decay. We fine-tune it for 160k iterations.

The comparison results are shown in \cref{tab:my-table}. Img2Vec achieves $50.7\%$ mIoU, surpassing all the other methods by a large margin. The results on the above two datasets demonstrate the transfer learning effectiveness of our method.

\vspace{-5pt}
\subsection{Ablation studies}
\vspace{-5pt}

In this section, we study the role of multi-block feature learning and global semantic learning. We use ViT-B as the backbone and evaluate ImageNet-1K top-1 accuracy.

\textbf{Multi-block feature learning.}
In this experiment, we trained the models for $1600$ epochs and did not adopt global semantic learning. To verify the module's effectiveness, we adopt MAE ViT-B, MAE ViT-L, and DINO ResNet-50 as the teacher, respectively. The results are shown in \cref{tab:multi-block}. When using MAE ViT as the teacher, multi-block feature learning promotes the fine-tuning accuracy significantly. With DINO ResNet-50 being the teacher, multi-block feature learning also works.

\begin{table}
  \centering
  \begin{tabular}{@{}lcc@{}}
    \toprule
    Teacher & w/o multi-block & w/ multi-block \\
    \midrule
    MAE ViT-Base   & 84.30\% & 84.38\% \\
    MAE ViT-Large  & 84.40\% & 84.72\% \\
    DINO ResNet-50 & 84.98\% & 85.02\% \\
    \bottomrule
  \end{tabular}
  \caption{\textbf{Multi-block feature learning with various teachers.} It works on both ViT and ResNet teachers.}
  \label{tab:multi-block}
  \vspace{-10pt}
\end{table}

\begin{figure}
  \centering
  \includegraphics[width=0.98\linewidth]{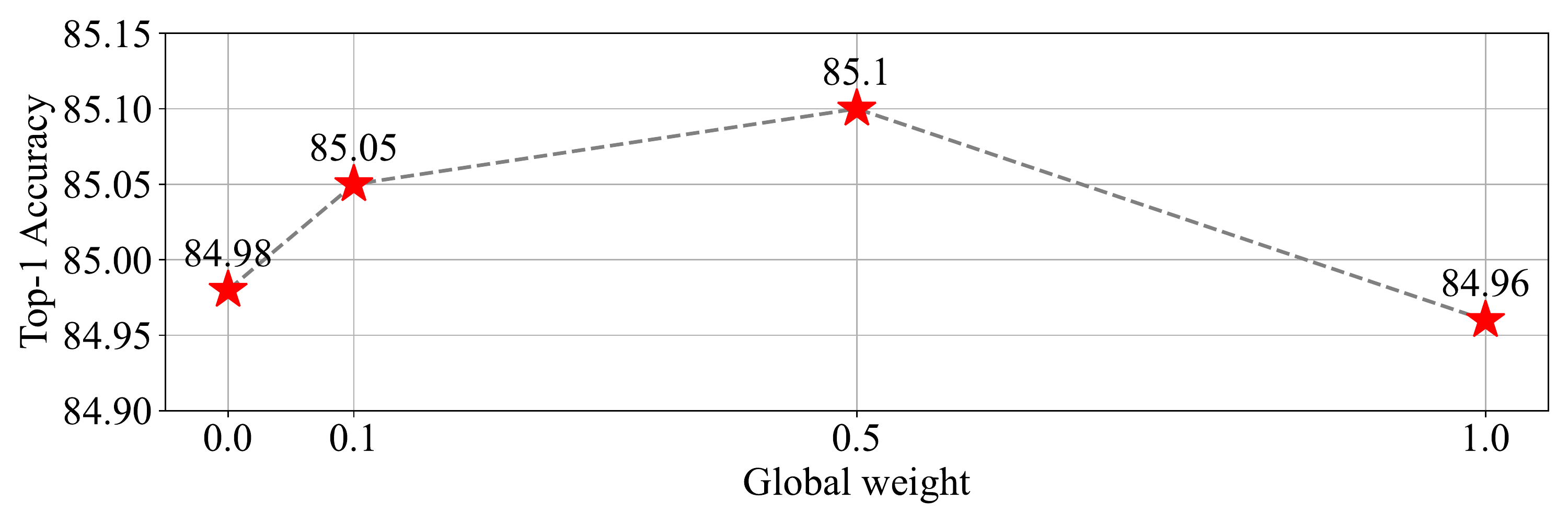}
  \caption{\textbf{Comparisons with different weights of the global loss.} We change the global weight from $0$ to $1$, and the top-1 accuracy shows a trend of first rising and then falling.}
  \label{fig:global_weight}
  \vspace{-15pt}
\end{figure}

\textbf{Loss weight of global semantic learning.} \cref{fig:global_weight} studies the influence of global loss weight for Img2Vec. We change the weight coefficient $\lambda$ from $0$ to $1$ and train the models for $1600$ epochs. Compared to the baseline ($\lambda = 0$), a relatively small weight improves $0.12\%$ accuracy absolutely, which shows that the global loss can benefit MIM from image-level supervision. However, a large weight ($\lambda=1$) will distract the pre-training and reduce the impact of MIM, thus getting a weaker result. Therefore, we set $\lambda = 0.5$ in Img2Vec.

\begin{table}[t]
  \centering
  \begin{tabular}{@{}cc||cc@{}}
    \toprule
    Projector head   & Top-1   & Loss         & Top-1    \\ \midrule
    2-layer MLP & 84.7\%  & Smooth $L_1$ & 84.7\%   \\
    Transformer & 84.0\%  & InfoNCE      & 83.7\%   \\  \bottomrule
  \end{tabular}
  \vspace{-5pt}
  \caption{\textbf{Left: Projector design.} We compare two different projector heads for Img2Vec. A simple 2-layer MLP projector is sufficient for global feature extracting. \textbf{Right: Loss type.} We compare two types of the global loss: smooth $L_1$ loss and InfoNCE loss. The smooth $L_1$ loss is better.}
  \label{tab:projector_loss}
  \vspace{-15pt}
\end{table}

\textbf{Projector head of global semantic learning.} We compare two projector structures in \cref{tab:projector_loss}. The 2-layer MLP projector head is simply stacked by two fully-connected layers and one activation layer ReLU\cite{nair2010rectified}. And the transformer projector head consists of two standard transformer blocks and one fully-connected layer. We conduct experiments on ViT-B with two types of projector and being pre-trained with $400$ epochs. 
Although transformer block projector increases the computational complexity, it degrades $0.7\%$ top-1 accuracy absolutely compared to MLP projector. We speculate that the heavier projector withdraws the capability from the pre-training encoder, and that a simple MLP is sufficient for global semantic learning. In Img2Vec, we use MLP as the projector head.

\textbf{Loss type of global semantic learning.} In \cref{tab:projector_loss}, we compare two loss types. Models are trained for $400$ epochs. The smooth $L_1$\cite{girshick2015fast} loss is frequently used in reconstruction tasks, while the InfoNCE\cite{oord2018representation} loss is a typical contrastive loss.
The augmentation between two views in MIM is inapparent, which eases the discrimination of positive pairs for InfoNCE loss, making the result worse than that of smooth $L_1$ loss by $1.0\%$ absolutely. Thus, we choose the smooth $L_1$ loss as the global loss.

\vspace{-5pt}

\section{Conclusions} \label{sec:conclusion}

Previous masked image modeling works mainly focus on the framework design\cite{chen2022context, dong2022bootstrapped, he2022masked}, masking strategy\cite{he2022masked, li2022semmae}, or combining with contrastive learning\cite{huang2022contrastive}. A few researches study the learning targets of MIM. MaskFeat\cite{wei2022masked} uses more effective low-level features like HOG\cite{dalal2005histograms} as its targets, whereas it still contains less semantic representations than deep features. MVP\cite{wei2022mvp} uses the multi-modality CLIP\cite{radford2021learning} model as a feature extractor, which leaks additional data implicitly during pre-training and restricts scaling up with richer visual modality data.

Different from previous works, in this paper, we study the property of MIM teachers for the first time. We first introduce a simple framework that uses the targets extracted from a trained model instead of an EMA model (\eg, data2vec\cite{baevski2022data2vec}) or a feature codebook (\eg, BEiT\cite{bao2021beit}). We find that a larger teacher model does not consistently lead to better student performance. Therefore, we introduce token diversity to evaluate teacher models and figure out that higher token-diversity helps MIM learning more forcefully. Detailed analyses are presented in \cref{sec:32}. When using a high token-diversity model, DINO\cite{caron2021emerging} ResNet-50, Img2Vec achieves $85.1\%$ top-1 accuracy on ImageNet\cite{russakovsky2015imagenet} with ViT-B\cite{dosovitskiy2020image}, outperforming all the competitors under the fair settings. Img2Vec also achieves state-of-the-art results on other downstream tasks.

A novel perspective, token diversity, is introduced to evaluate teacher models. However, other aspects of teachers may also impact, which calls for further investigations. And how to train a teacher model with higher token-diversity is also worthy of more studies.

\nocite{langley00}

\bibliography{egbib}

\begin{thebibliography}{54}
\providecommand{\natexlab}[1]{#1}
\providecommand{\url}[1]{\texttt{#1}}
\expandafter\ifx\csname urlstyle\endcsname\relax
  \providecommand{\doi}[1]{doi: #1}\else
  \providecommand{\doi}{doi: \begingroup \urlstyle{rm}\Url}\fi

\bibitem[Baevski et~al.(2022)Baevski, Hsu, Xu, Babu, Gu, and
  Auli]{baevski2022data2vec}
Baevski, A., Hsu, W., Xu, Q., Babu, A., Gu, J., and Auli, M.
\newblock data2vec: {A} general framework for self-supervised learning in
  speech, vision and language.
\newblock \emph{arXiv preprint arXiv:2202.03555}, 2022.

\bibitem[Bai et~al.(2022)Bai, Yuan, Xia, Yan, Li, and Liu]{bai2022improving}
Bai, J., Yuan, L., Xia, S.-T., Yan, S., Li, Z., and Liu, W.
\newblock Improving vision transformers by revisiting high-frequency
  components.
\newblock In \emph{Computer Vision--ECCV 2022: 17th European Conference, Tel
  Aviv, Israel, October 23--27, 2022, Proceedings, Part XXIV}, pp.\  1--18.
  Springer, 2022.

\bibitem[Bao et~al.(2022)Bao, Dong, Piao, and Wei]{bao2021beit}
Bao, H., Dong, L., Piao, S., and Wei, F.
\newblock {BE}it: {BERT} pre-training of image transformers.
\newblock In \emph{International Conference on Learning Representations}, 2022.

\bibitem[Cai \& Vasconcelos(2018)Cai and Vasconcelos]{cai2019cascade}
Cai, Z. and Vasconcelos, N.
\newblock Cascade r-cnn: Delving into high quality object detection.
\newblock In \emph{Proceedings of the IEEE Conference on Computer Vision and
  Pattern Recognition}, 2018.

\bibitem[Caron et~al.(2021)Caron, Touvron, Misra, J\'egou, Mairal, Bojanowski,
  and Joulin]{caron2021emerging}
Caron, M., Touvron, H., Misra, I., J\'egou, H., Mairal, J., Bojanowski, P., and
  Joulin, A.
\newblock Emerging properties in self-supervised vision transformers.
\newblock In \emph{Proceedings of the IEEE/CVF International Conference on
  Computer Vision}, pp.\  9650--9660, 2021.

\bibitem[Chen et~al.(2020{\natexlab{a}})Chen, Radford, Child, Wu, Jun, Luan,
  and Sutskever]{chen2020generative}
Chen, M., Radford, A., Child, R., Wu, J., Jun, H., Luan, D., and Sutskever, I.
\newblock Generative pretraining from pixels.
\newblock In \emph{International Conference on Machine Learning}, pp.\
  1691--1703. PMLR, 2020{\natexlab{a}}.

\bibitem[Chen et~al.(2020{\natexlab{b}})Chen, Kornblith, Norouzi, and
  Hinton]{chen2020simple}
Chen, T., Kornblith, S., Norouzi, M., and Hinton, G.
\newblock A simple framework for contrastive learning of visual
  representations.
\newblock In \emph{Proceedings of the 37th International Conference on Machine
  Learning}, pp.\  1597--1607, 2020{\natexlab{b}}.

\bibitem[Chen \& He(2021)Chen and He]{chen2021exploring}
Chen, X. and He, K.
\newblock Exploring simple siamese representation learning.
\newblock In \emph{Proceedings of the IEEE/CVF Conference on Computer Vision
  and Pattern Recognition}, pp.\  15750--15758, 2021.

\bibitem[Chen et~al.(2021)Chen, Xie, and He]{chen2021empirical}
Chen, X., Xie, S., and He, K.
\newblock An empirical study of training self-supervised vision transformers.
\newblock In \emph{Proceedings of the IEEE/CVF International Conference on
  Computer Vision}, pp.\  9640--9649, 2021.

\bibitem[Chen et~al.(2022)Chen, Ding, Wang, Xin, Mo, Wang, Han, Luo, Zeng, and
  Wang]{chen2022context}
Chen, X., Ding, M., Wang, X., Xin, Y., Mo, S., Wang, Y., Han, S., Luo, P.,
  Zeng, G., and Wang, J.
\newblock Context autoencoder for self-supervised representation learning.
\newblock \emph{arXiv preprint arXiv:2202.03026}, 2022.

\bibitem[Clark et~al.(2020)Clark, Luong, Le, and Manning]{clark2020electra}
Clark, K., Luong, M.-T., Le, Q.~V., and Manning, C.~D.
\newblock {ELECTRA}: Pre-training text encoders as discriminators rather than
  generators.
\newblock In \emph{International Conference on Learning Representations}, 2020.

\bibitem[Dalal \& Triggs(2005)Dalal and Triggs]{dalal2005histograms}
Dalal, N. and Triggs, B.
\newblock Histograms of oriented gradients for human detection.
\newblock In \emph{IEEE Computer Society Conference on Computer Vision and
  Pattern Recognition}, volume~1, pp.\  886--893, 2005.

\bibitem[Devlin et~al.(2019)Devlin, Chang, Lee, and Toutanova]{devlin2018bert}
Devlin, J., Chang, M., Lee, K., and Toutanova, K.
\newblock {BERT:} pre-training of deep bidirectional transformers for language
  understanding.
\newblock In \emph{Proceedings of the Conference of the North American Chapter
  of the Association for Computational Linguistics}, 2019.

\bibitem[Dong et~al.(2022)Dong, Bao, Zhang, Chen, Zhang, Yuan, Chen, Wen, and
  Yu]{dong2022bootstrapped}
Dong, X., Bao, J., Zhang, T., Chen, D., Zhang, W., Yuan, L., Chen, D., Wen, F.,
  and Yu, N.
\newblock Bootstrapped masked autoencoders for vision bert pretraining.
\newblock \emph{Proceedings of the European Conference on Computer Vision},
  2022.

\bibitem[Dong et~al.(2023)Dong, Bao, Zhang, Chen, Zhang, Yuan, Chen, Wen, and
  Peco]{dong2023peco}
Dong, X., Bao, J., Zhang, T., Chen, D., Zhang, W., Yuan, L., Chen, D., Wen, F.,
  and Peco, N.~Y.
\newblock Peco: Perceptual codebook for bert pre-training of vision
  transformers.
\newblock \emph{Association for the Advancement of Artificial Intelligence},
  2023.

\bibitem[Dosovitskiy et~al.(2021)Dosovitskiy, Beyer, Kolesnikov, Weissenborn,
  Zhai, Unterthiner, Dehghani, Minderer, Heigold, Gelly, Uszkoreit, and
  Houlsby]{dosovitskiy2020image}
Dosovitskiy, A., Beyer, L., Kolesnikov, A., Weissenborn, D., Zhai, X.,
  Unterthiner, T., Dehghani, M., Minderer, M., Heigold, G., Gelly, S.,
  Uszkoreit, J., and Houlsby, N.
\newblock An image is worth 16x16 words: Transformers for image recognition at
  scale.
\newblock In \emph{International Conference on Learning Representations}, 2021.

\bibitem[Girshick(2015)]{girshick2015fast}
Girshick, R.
\newblock Fast r-cnn.
\newblock In \emph{Proceedings of the IEEE international conference on computer
  vision}, pp.\  1440--1448, 2015.

\bibitem[Glorot \& Bengio(2010)Glorot and Bengio]{glorot2010understanding}
Glorot, X. and Bengio, Y.
\newblock Understanding the difficulty of training deep feedforward neural
  networks.
\newblock In \emph{Proceedings of the thirteenth international conference on
  artificial intelligence and statistics}, pp.\  249--256. JMLR Workshop and
  Conference Proceedings, 2010.

\bibitem[Goyal et~al.(2017)Goyal, Doll{\'a}r, Girshick, Noordhuis, Wesolowski,
  Kyrola, Tulloch, Jia, and He]{goyal2017accurate}
Goyal, P., Doll{\'a}r, P., Girshick, R., Noordhuis, P., Wesolowski, L., Kyrola,
  A., Tulloch, A., Jia, Y., and He, K.
\newblock Accurate, large minibatch sgd: Training imagenet in 1 hour.
\newblock \emph{arXiv preprint arXiv:1706.02677}, 2017.

\bibitem[Goyal et~al.(2019)Goyal, Mahajan, Gupta, and Misra]{goyal2019scaling}
Goyal, P., Mahajan, D., Gupta, A., and Misra, I.
\newblock Scaling and benchmarking self-supervised visual representation
  learning.
\newblock In \emph{Proceedings of the IEEE/CVF International Conference on
  computer vision}, pp.\  6391--6400, 2019.

\bibitem[Grill et~al.(2020)Grill, Strub, Altch\'{e}, Tallec, Richemond,
  Buchatskaya, Doersch, Avila~Pires, Guo, Gheshlaghi~Azar, Piot, kavukcuoglu,
  Munos, and Valko]{grill2020bootstrap}
Grill, J.-B., Strub, F., Altch\'{e}, F., Tallec, C., Richemond, P.,
  Buchatskaya, E., Doersch, C., Avila~Pires, B., Guo, Z., Gheshlaghi~Azar, M.,
  Piot, B., kavukcuoglu, k., Munos, R., and Valko, M.
\newblock Bootstrap your own latent - {A} new approach to self-supervised
  learning.
\newblock In \emph{Advances in Neural Information Processing Systems}, pp.\
  21271--21284, 2020.

\bibitem[He et~al.(2016)He, Zhang, Ren, and Sun]{he2016deep}
He, K., Zhang, X., Ren, S., and Sun, J.
\newblock Deep residual learning for image recognition.
\newblock In \emph{Proceedings of the IEEE conference on Computer Vision and
  Pattern Recognition}, pp.\  770--778, 2016.

\bibitem[He et~al.(2017)He, Gkioxari, Dollar, and Girshick]{he2017mask}
He, K., Gkioxari, G., Dollar, P., and Girshick, R.
\newblock Mask r-cnn.
\newblock In \emph{Proceedings of the IEEE International Conference on Computer
  Vision}, 2017.

\bibitem[He et~al.(2020)He, Fan, Wu, Xie, and Girshick]{he2020momentum}
He, K., Fan, H., Wu, Y., Xie, S., and Girshick, R.
\newblock Momentum contrast for unsupervised visual representation learning.
\newblock In \emph{Proceedings of the IEEE/CVF conference on Computer Vision
  and Pattern Recognition}, pp.\  9729--9738, 2020.

\bibitem[He et~al.(2022)He, Chen, Xie, Li, Doll{\'a}r, and
  Girshick]{he2022masked}
He, K., Chen, X., Xie, S., Li, Y., Doll{\'a}r, P., and Girshick, R.
\newblock Masked autoencoders are scalable vision learners.
\newblock In \emph{Proceedings of the IEEE/CVF Conference on Computer Vision
  and Pattern Recognition}, pp.\  16000--16009, 2022.

\bibitem[Huang et~al.(2016)Huang, Sun, Liu, Sedra, and
  Weinberger]{huang2016deep}
Huang, G., Sun, Y., Liu, Z., Sedra, D., and Weinberger, K.~Q.
\newblock Deep networks with stochastic depth.
\newblock In \emph{Proceedings of the European Conference on Computer Vision},
  pp.\  646--661, 2016.

\bibitem[Huang et~al.(2022)Huang, Jin, Lu, Hou, Cheng, Fu, Shen, and
  Feng]{huang2022contrastive}
Huang, Z., Jin, X., Lu, C., Hou, Q., Cheng, M.-M., Fu, D., Shen, X., and Feng,
  J.
\newblock Contrastive masked autoencoders are stronger vision learners.
\newblock \emph{arXiv preprint arXiv:2207.13532}, 2022.

\bibitem[Kakogeorgiou et~al.(2022)Kakogeorgiou, Gidaris, Psomas, Avrithis,
  Bursuc, Karantzalos, and Komodakis]{kakogeorgiou2022hide}
Kakogeorgiou, I., Gidaris, S., Psomas, B., Avrithis, Y., Bursuc, A.,
  Karantzalos, K., and Komodakis, N.
\newblock What to hide from your students: Attention-guided masked image
  modeling.
\newblock In \emph{Proceedings of the European Conference on Computer Vision},
  pp.\  300--318, 2022.

\bibitem[Komodakis \& Gidaris(2018)Komodakis and
  Gidaris]{komodakis2018unsupervised}
Komodakis, N. and Gidaris, S.
\newblock Unsupervised representation learning by predicting image rotations.
\newblock In \emph{International Conference on Learning Representations}, 2018.

\bibitem[Li et~al.(2022)Li, Zheng, Liu, Wang, Su, and Zheng]{li2022semmae}
Li, G., Zheng, H., Liu, D., Wang, C., Su, B., and Zheng, C.
\newblock Semmae: Semantic-guided masking for learning masked autoencoders.
\newblock \emph{Advances in Neural Information Processing Systems}, 2022.

\bibitem[Lin et~al.(2014)Lin, Maire, Belongie, Hays, Perona, Ramanan,
  Doll{\'a}r, and Zitnick]{lin2014microsoft}
Lin, T.-Y., Maire, M., Belongie, S., Hays, J., Perona, P., Ramanan, D.,
  Doll{\'a}r, P., and Zitnick, C.~L.
\newblock Microsoft coco: Common objects in context.
\newblock In \emph{Proceedings of the European Conference on Computer Vision},
  pp.\  740--755, 2014.

\bibitem[Lin et~al.(2017)Lin, Doll{\'a}r, Girshick, He, Hariharan, and
  Belongie]{lin2017feature}
Lin, T.-Y., Doll{\'a}r, P., Girshick, R., He, K., Hariharan, B., and Belongie,
  S.
\newblock Feature pyramid networks for object detection.
\newblock In \emph{Proceedings of the IEEE conference on Computer Vision and
  Pattern Recognition}, pp.\  2117--2125, 2017.

\bibitem[Loshchilov \& Hutter(2017{\natexlab{a}})Loshchilov and
  Hutter]{loshchilov2016sgdr}
Loshchilov, I. and Hutter, F.
\newblock {SGDR:} stochastic gradient descent with warm restarts.
\newblock In \emph{International Conference on Learning Representations},
  2017{\natexlab{a}}.

\bibitem[Loshchilov \& Hutter(2017{\natexlab{b}})Loshchilov and
  Hutter]{loshchilov2017sgdr}
Loshchilov, I. and Hutter, F.
\newblock Sgdr: Stochastic gradient descent with warm restarts.
\newblock \emph{International Conference on Learning Representations},
  2017{\natexlab{b}}.

\bibitem[Loshchilov \& Hutter(2019{\natexlab{a}})Loshchilov and
  Hutter]{loshchilov2018decoupled}
Loshchilov, I. and Hutter, F.
\newblock Decoupled weight decay regularization.
\newblock In \emph{International Conference on Learning Representations},
  2019{\natexlab{a}}.

\bibitem[Loshchilov \& Hutter(2019{\natexlab{b}})Loshchilov and
  Hutter]{loshchilov2019decoupled}
Loshchilov, I. and Hutter, F.
\newblock Decoupled weight decay regularization.
\newblock \emph{International Conference on Learning Representations},
  2019{\natexlab{b}}.

\bibitem[Nair \& Hinton(2010)Nair and Hinton]{nair2010rectified}
Nair, V. and Hinton, G.~E.
\newblock Rectified linear units improve restricted boltzmann machines.
\newblock In \emph{{Proceedings of the International Conference on Machine
  Learning}}, pp.\  807--814, 2010.

\bibitem[Noroozi \& Favaro(2016)Noroozi and Favaro]{noroozi2016unsupervised}
Noroozi, M. and Favaro, P.
\newblock Unsupervised learning of visual representations by solving jigsaw
  puzzles.
\newblock In \emph{Proceedings of the European Conference on Computer Vision},
  pp.\  69--84, 2016.

\bibitem[Oord et~al.(2018)Oord, Li, and Vinyals]{oord2018representation}
Oord, A. v.~d., Li, Y., and Vinyals, O.
\newblock Representation learning with contrastive predictive coding.
\newblock \emph{arXiv preprint arXiv:1807.03748}, 2018.

\bibitem[Radford et~al.(2021)Radford, Kim, Hallacy, Ramesh, Goh, Agarwal,
  Sastry, Askell, Mishkin, Clark, et~al.]{radford2021learning}
Radford, A., Kim, J.~W., Hallacy, C., Ramesh, A., Goh, G., Agarwal, S., Sastry,
  G., Askell, A., Mishkin, P., Clark, J., et~al.
\newblock Learning transferable visual models from natural language
  supervision.
\newblock In \emph{Proceedings of the International Conference on Machine
  Learning}, pp.\  8748--8763, 2021.

\bibitem[Raffel et~al.(2020)Raffel, Shazeer, Roberts, Lee, Narang, Matena,
  Zhou, Li, Liu, et~al.]{raffel2020exploring}
Raffel, C., Shazeer, N., Roberts, A., Lee, K., Narang, S., Matena, M., Zhou,
  Y., Li, W., Liu, P.~J., et~al.
\newblock Exploring the limits of transfer learning with a unified text-to-text
  transformer.
\newblock \emph{Journal of Machine Learning Research}, 21\penalty0
  (140):\penalty0 1--67, 2020.

\bibitem[Russakovsky et~al.(2015)Russakovsky, Deng, Su, Krause, Satheesh, Ma,
  Huang, Karpathy, Khosla, Bernstein, et~al.]{russakovsky2015imagenet}
Russakovsky, O., Deng, J., Su, H., Krause, J., Satheesh, S., Ma, S., Huang, Z.,
  Karpathy, A., Khosla, A., Bernstein, M., et~al.
\newblock Imagenet large scale visual recognition challenge.
\newblock \emph{International Journal of Computer Vision}, 115\penalty0
  (3):\penalty0 211--252, 2015.

\bibitem[Shaw et~al.(2018)Shaw, Uszkoreit, and Vaswani]{shaw2018self}
Shaw, P., Uszkoreit, J., and Vaswani, A.
\newblock Self-attention with relative position representations.
\newblock In \emph{Proceedings of the 2018 Conference of the North American
  Chapter of the Association for Computational Linguistics}, pp.\  464--468,
  2018.

\bibitem[Su et~al.(2020)Su, Zhu, Cao, Li, Lu, Wei, and Dai]{su2019vl}
Su, W., Zhu, X., Cao, Y., Li, B., Lu, L., Wei, F., and Dai, J.
\newblock {VL-BERT:} pre-training of generic visual-linguistic representations.
\newblock In \emph{International Conference on Learning Representations}, 2020.

\bibitem[Szegedy et~al.(2016)Szegedy, Vanhoucke, Ioffe, Shlens, and
  Wojna]{szegedy2016rethinking}
Szegedy, C., Vanhoucke, V., Ioffe, S., Shlens, J., and Wojna, Z.
\newblock Rethinking the inception architecture for computer vision.
\newblock In \emph{Proceedings of the IEEE conference on Computer Vision and
  Pattern Recognition}, pp.\  2818--2826, 2016.

\bibitem[Wei et~al.(2022{\natexlab{a}})Wei, Fan, Xie, Wu, Yuille, and
  Feichtenhofer]{wei2022masked}
Wei, C., Fan, H., Xie, S., Wu, C.-Y., Yuille, A., and Feichtenhofer, C.
\newblock Masked feature prediction for self-supervised visual pre-training.
\newblock In \emph{Proceedings of the IEEE/CVF Conference on Computer Vision
  and Pattern Recognition}, pp.\  14668--14678, 2022{\natexlab{a}}.

\bibitem[Wei et~al.(2022{\natexlab{b}})Wei, Xie, Zhou, Li, and
  Tian]{wei2022mvp}
Wei, L., Xie, L., Zhou, W., Li, H., and Tian, Q.
\newblock Mvp: Multimodality-guided visual pre-training.
\newblock \emph{arXiv preprint arXiv:2203.05175}, 2022{\natexlab{b}}.

\bibitem[Xiao et~al.(2018)Xiao, Liu, Zhou, Jiang, and Sun]{xiao2018unified}
Xiao, T., Liu, Y., Zhou, B., Jiang, Y., and Sun, J.
\newblock Unified perceptual parsing for scene understanding.
\newblock In \emph{Proceedings of the European Conference on Computer Vision},
  pp.\  418--434, 2018.

\bibitem[Xie et~al.(2022)Xie, Zhang, Cao, Lin, Bao, Yao, Dai, and
  Hu]{xie2022simmim}
Xie, Z., Zhang, Z., Cao, Y., Lin, Y., Bao, J., Yao, Z., Dai, Q., and Hu, H.
\newblock Simmim: A simple framework for masked image modeling.
\newblock In \emph{Proceedings of the IEEE/CVF Conference on Computer Vision
  and Pattern Recognition}, pp.\  9653--9663, 2022.

\bibitem[Yun et~al.(2019)Yun, Han, Oh, Chun, Choe, and Yoo]{yun2019cutmix}
Yun, S., Han, D., Oh, S.~J., Chun, S., Choe, J., and Yoo, Y.
\newblock Cutmix: Regularization strategy to train strong classifiers with
  localizable features.
\newblock In \emph{Proceedings of the IEEE/CVF international conference on
  computer vision}, pp.\  6023--6032, 2019.

\bibitem[Zhang et~al.(2018)Zhang, Ciss{\'{e}}, Dauphin, and
  Lopez{-}Paz]{zhang2017mixup}
Zhang, H., Ciss{\'{e}}, M., Dauphin, Y.~N., and Lopez{-}Paz, D.
\newblock mixup: Beyond empirical risk minimization.
\newblock In \emph{International Conference on Learning Representations}, 2018.

\bibitem[Zhang et~al.(2022)Zhang, Wang, and Wang]{zhang2022mask}
Zhang, Q., Wang, Y., and Wang, Y.
\newblock How mask matters: Towards theoretical understandings of masked
  autoencoders.
\newblock In \emph{Advances in Neural Information Processing Systems}, 2022.

\bibitem[Zhou et~al.(2019)Zhou, Zhao, Puig, Xiao, Fidler, Barriuso, and
  Torralba]{zhou2019semantic}
Zhou, B., Zhao, H., Puig, X., Xiao, T., Fidler, S., Barriuso, A., and Torralba,
  A.
\newblock Semantic understanding of scenes through the ade20k dataset.
\newblock \emph{International Journal of Computer Vision}, 127\penalty0
  (3):\penalty0 302--321, 2019.

\bibitem[Zhou et~al.(2022)Zhou, Wei, Wang, Shen, Xie, Yuille, and
  Kong]{zhou2021ibot}
Zhou, J., Wei, C., Wang, H., Shen, W., Xie, C., Yuille, A., and Kong, T.
\newblock Image {BERT} pre-training with online tokenizer.
\newblock In \emph{International Conference on Learning Representations}, 2022.

\end{thebibliography}
\bibliographystyle{icml2023}

\clearpage
\appendix
\section{Teacher Models}

We consider $5$ pre-trained models from $3$ representative self-supervised approaches for teacher comparisons. Models trained with additional data or labeled data are excluded (\eg,  CLIP\cite{radford2021learning}, fine-tuned ResNet\cite{he2016deep}).
Selected teachers for fair comparisons are:
\begin{itemize}
    \item \textbf{MAE\cite{he2022masked}}. The most representative pre-training method of masked image modeling. It optimizes the masked patches to reconstruct the image pixels and achieves stronger results than all contrastive learning methods. MAE trained models on ViT\cite{dosovitskiy2020image}, and we use ViT-B/16 and ViT-L/16 as teachers.
    
    \item \textbf{SimCLR\cite{chen2020simple}}. A simple yet effective framework of contrastive learning. It uses the same encoder to extract features from two augmented views of an image. All samples from the whole batch are collected to generate negative pairs for cross-entropy loss. SimCLR shows competitive results among contrastive learning methods. It provides pre-trained models on ResNet, and we use ResNet-50 for experiments.
    
    \item \textbf{DINO\cite{caron2021emerging}}. A powerful self-supervised method with self-distillation. It extracts the feature from the student model with a heavily augmented image and also obtains the features from the momentum-updated model with multiple other views. A [CLS] token is adopted for aggregating embeddings from all other tokens. The cross-entropy loss on features of positive pairs is used for optimizing. Different from other contrastive learning methods, DINO does not need negative pairs. It also achieves superb performance. We use ViT-B/16 and ResNet-50\cite{he2016deep} during teacher studies.
\end{itemize}

When taking these models as teachers, we initialize the model parameters with official checkpoints and freeze them during pre-training.


\section{Implementation Details of Img2Vec}

\subsection{ImageNet Experiments}

\textbf{Pre-training.} We list the pre-training settings in \cref{tab:pretrain}. Main settings follow the MAE\cite{he2022masked}. We use the standard ViT architecture without relative position embedding\cite{shaw2018self} and layer scaling\cite{bao2021beit}. All vision transformer blocks are initialized through Xavier Uniform\cite{glorot2010understanding}. We do not use stochastic drop path\cite{huang2016deep} on ViT-B and set drop path rate $0.1$ on ViT-L.

\textbf{Fine-tuning.} We follow the settings of MAE for end-to-end fine-tuning, except using a smaller learning rate. The detailed settings are presented in \cref{tab:finetune}.

\textbf{Linear probing.} We follow the settings of MAE for linear probing, all settings are unchanged.

\begin{table*}
\centering
    \begin{tabular}{@{}l|l@{}}
        Config                 & Value             \\ \midrule
        optimizer              & AdamW\cite{loshchilov2019decoupled} \\
        base learning rate     & $1.5\mathrm{e}{-4}$      \\
        weight decay           & $0.05$              \\
        optimizer momentum     & $\beta_1,\beta_2=0.9,0.95$ \cite{chen2020generative}  \\
        batch size             & $4096$              \\
        learning rate schedule & cosine decay\cite{loshchilov2017sgdr} \\
        warmup epochs\cite{goyal2017accurate} & $40$                \\
        augmentation           & RandomResizedCrop\cite{he2022masked} \\
        drop path rate\cite{huang2016deep} & $0$ (B), $0.1$ (L), $0.2$ (H)    \\
        pre-train epochs       & $1600$              \\
        teacher                & DINO ResNet-50    \\
        global loss weight     & $0.5$ (B, L), $0.1$ (H)     \\
        decoder depth          & $2$ (B), $4$ (L), $8$ (H)    \\
        patch size             & $16\times 16$ (B, L), $14 \times 14$ (H)    \\
        mask ratio             & $0.6$               \\
        mask block             & $32\times32$ \cite{xie2022simmim} \\
    \end{tabular}
    \caption{\textbf{Pre-training settings for ImageNet experiments.}}
    \label{tab:pretrain}
\end{table*}

\begin{table*}
\centering
    \begin{tabular}{@{}l|l@{}}
        Config                 & Value             \\ \midrule
        optimizer              & AdamW             \\
        base learning rate     & $1.25\mathrm{e}{-4}$  (B), $2\mathrm{e}{-4}$  (L,H)      \\
        weight decay           & $0.05$              \\
        layer-wise lr decay\cite{bao2021beit} & $0.65$ (B), $0.75$ (L, H) \\
        optimizer momentum     & $\beta_1,\beta_2=0.9,0.999$                   \\
        batch size             & $1024$              \\
        learning rate schedule & cosine decay      \\
        warmup epochs          & $5$                \\
        augmentation           & RandAug ($9$, $0.5$) \cite{he2022masked} \\
        drop path rate         & $0.1$ (B), $0.2$ (L), $0.3$ (H)    \\
        patch size             & $16\times 16$ (B, L), $14 \times 14$ (H)    \\
        label smoothing\cite{szegedy2016rethinking} & $0.1$ \\
        mixup\cite{zhang2017mixup} & $0.8$ \\
        cutmix\cite{yun2019cutmix} & $1.0$ \\
    \end{tabular}
    \caption{\textbf{Fine-tuning settings for ImageNet experiments.}}
    \label{tab:finetune}
\end{table*}

\subsection{COCO and ADE20K Experiments}

We transfer the model pre-trained by our Img2Vec on various downstream tasks. We use Mask R-CNN\cite{he2017mask} for object detection and instance segmentation on COCO\cite{lin2014microsoft} and use UperNet\cite{xiao2018unified} for semantic segmentation on ADE20K\cite{zhou2019semantic}. All settings keep the same with MAE, except a smaller learning rate ($4\mathrm{e}{-5}$ on COCO and $2.5\mathrm{e}{-4}$ on ADE20K).

\section{More Visualization}

\begin{figure*}
    \centering
        \begin{center}
            \quad\quad Input\qquad\quad\quad MAE ViT-B\qquad\quad MAE ViT-L\qquad SimCLR RN-50\qquad DINO ViT-B\qquad DINO RN-50\quad
        \end{center}
    \includegraphics[width=0.96\textwidth]{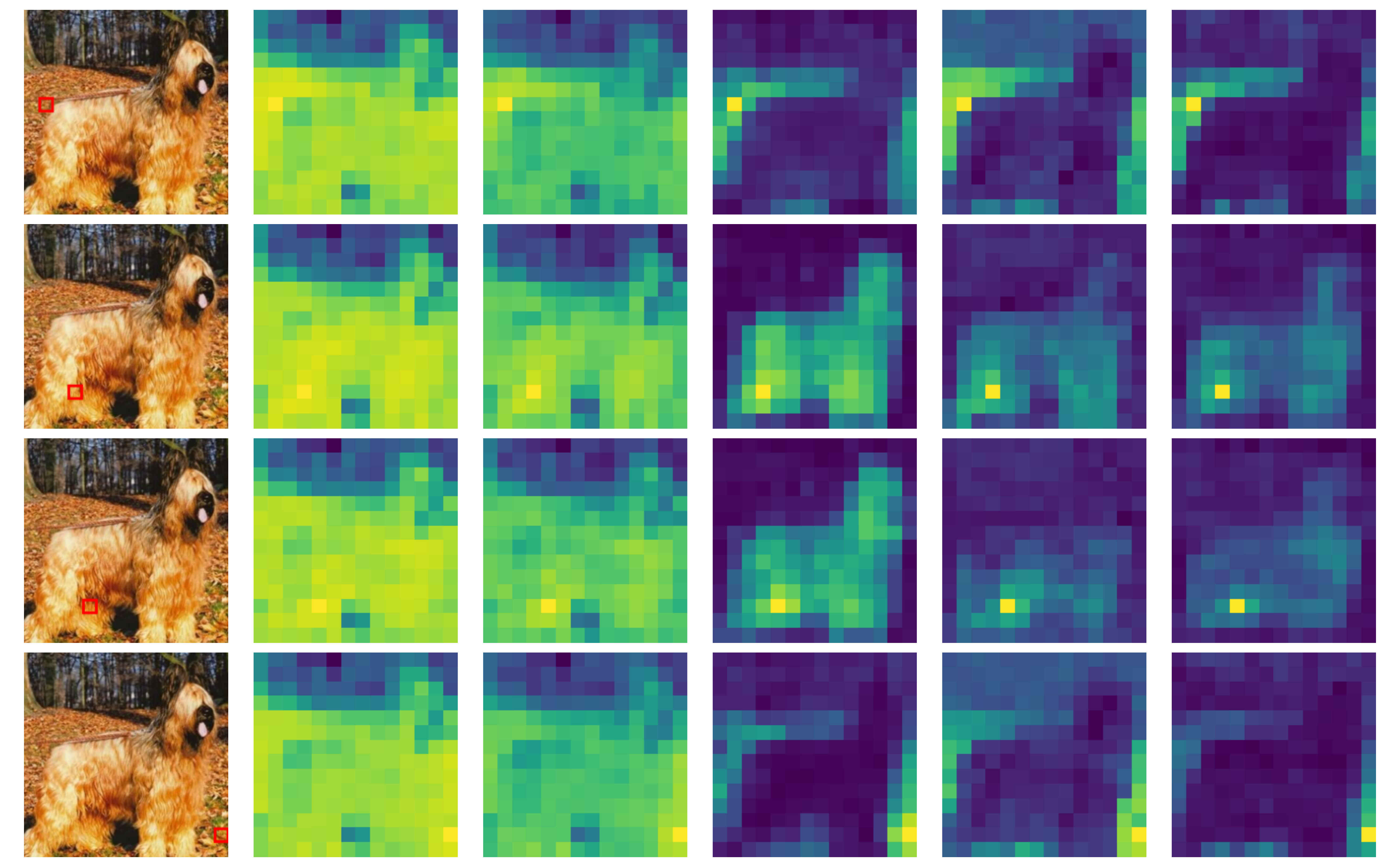}
    \includegraphics[width=0.96\textwidth]{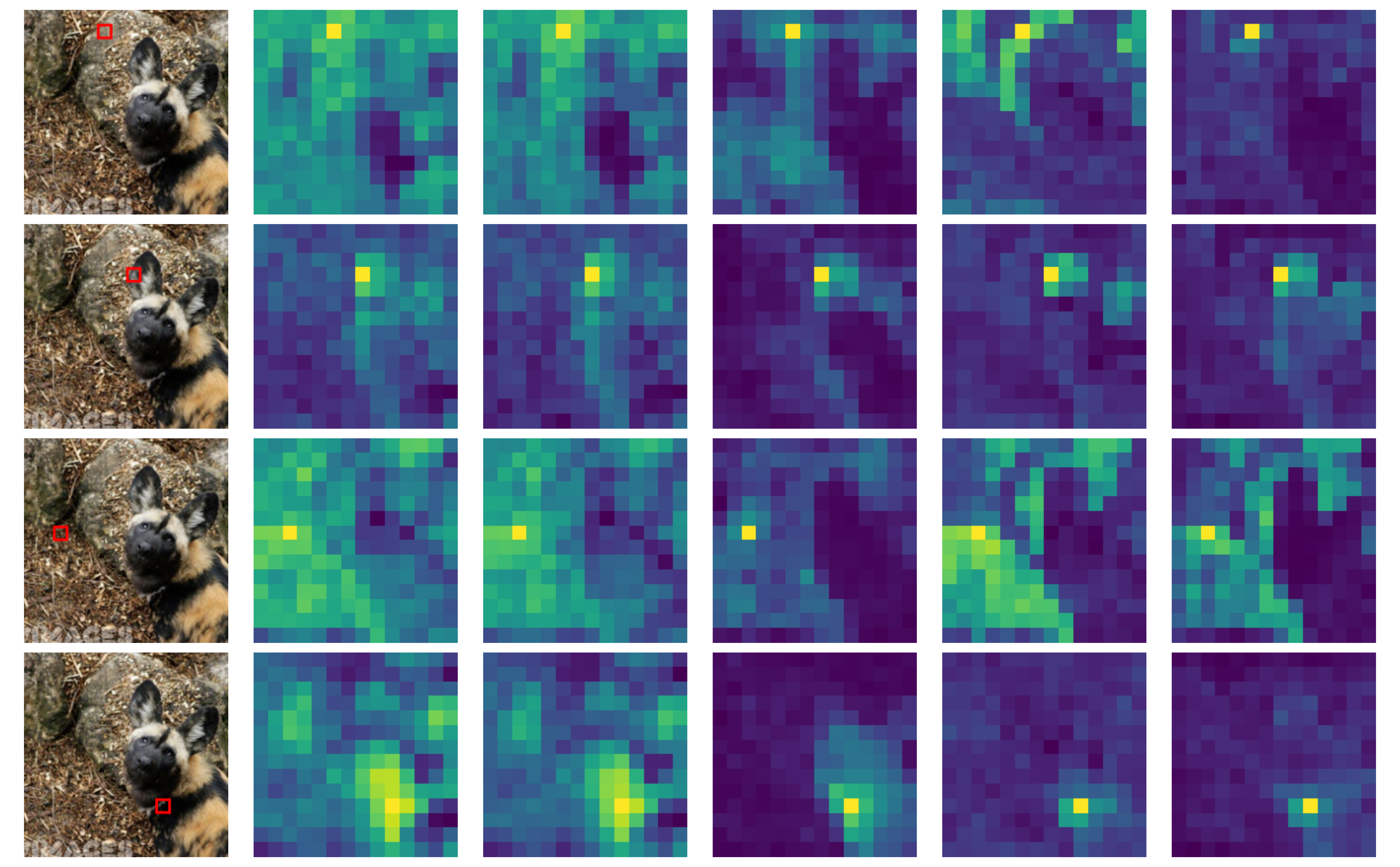}
    \caption{\textbf{Visualization of targets similarity from different teachers.} The area with a red box in the first column is the \textit{query} patch. The right 5 columns represent similarity heat-maps of different teachers. Each grid indicates its similarity with the \textit{query}.}
    \label{fig:appendix_show1}
\end{figure*}

\begin{figure*}
    \centering
        \begin{center}
            \quad\quad Input\qquad\quad\quad MAE ViT-B\qquad\quad MAE ViT-L\qquad SimCLR RN-50\qquad DINO ViT-B\qquad DINO RN-50\quad
        \end{center}
    \includegraphics[width=0.96\textwidth]{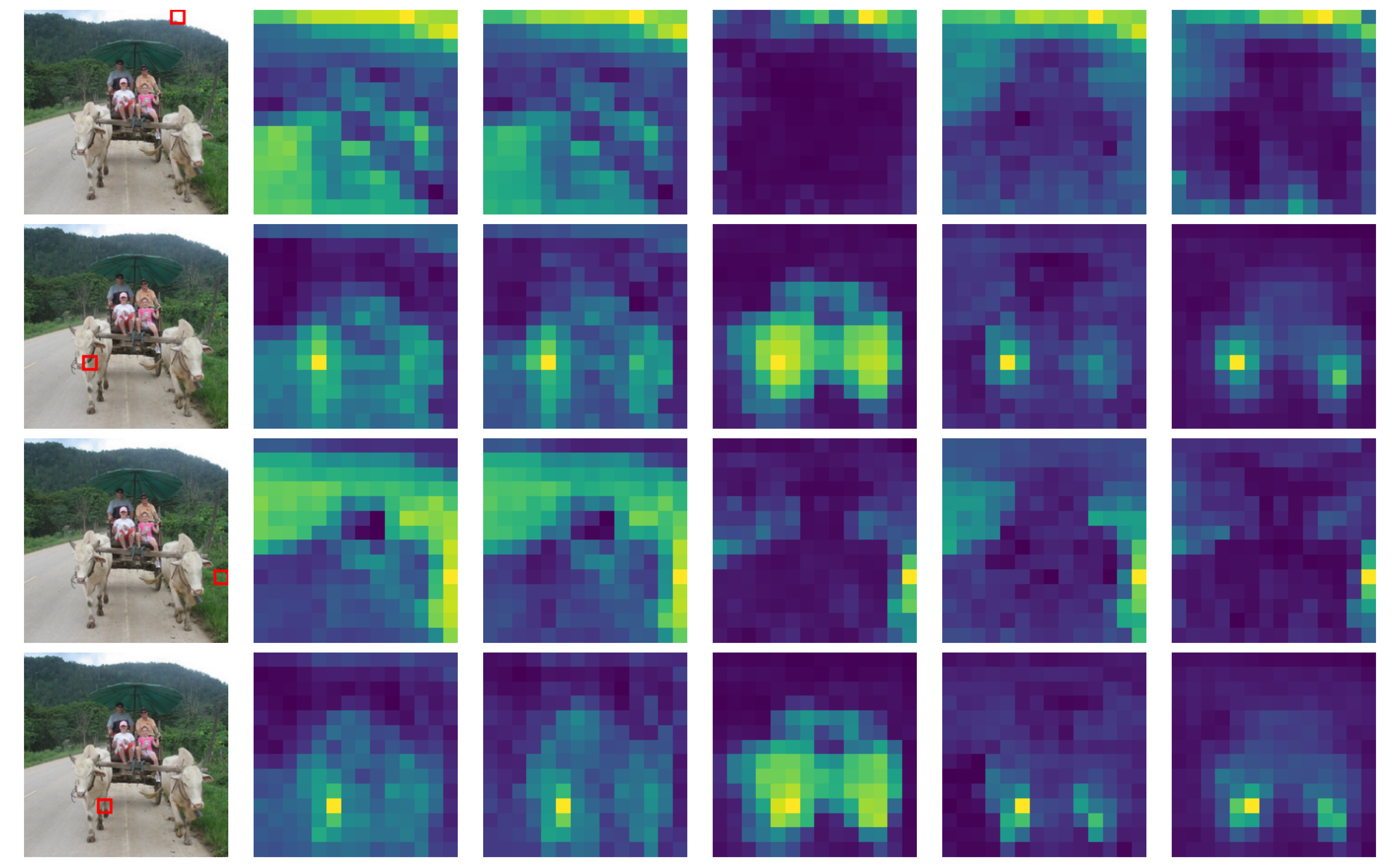}
    \includegraphics[width=0.96\textwidth]{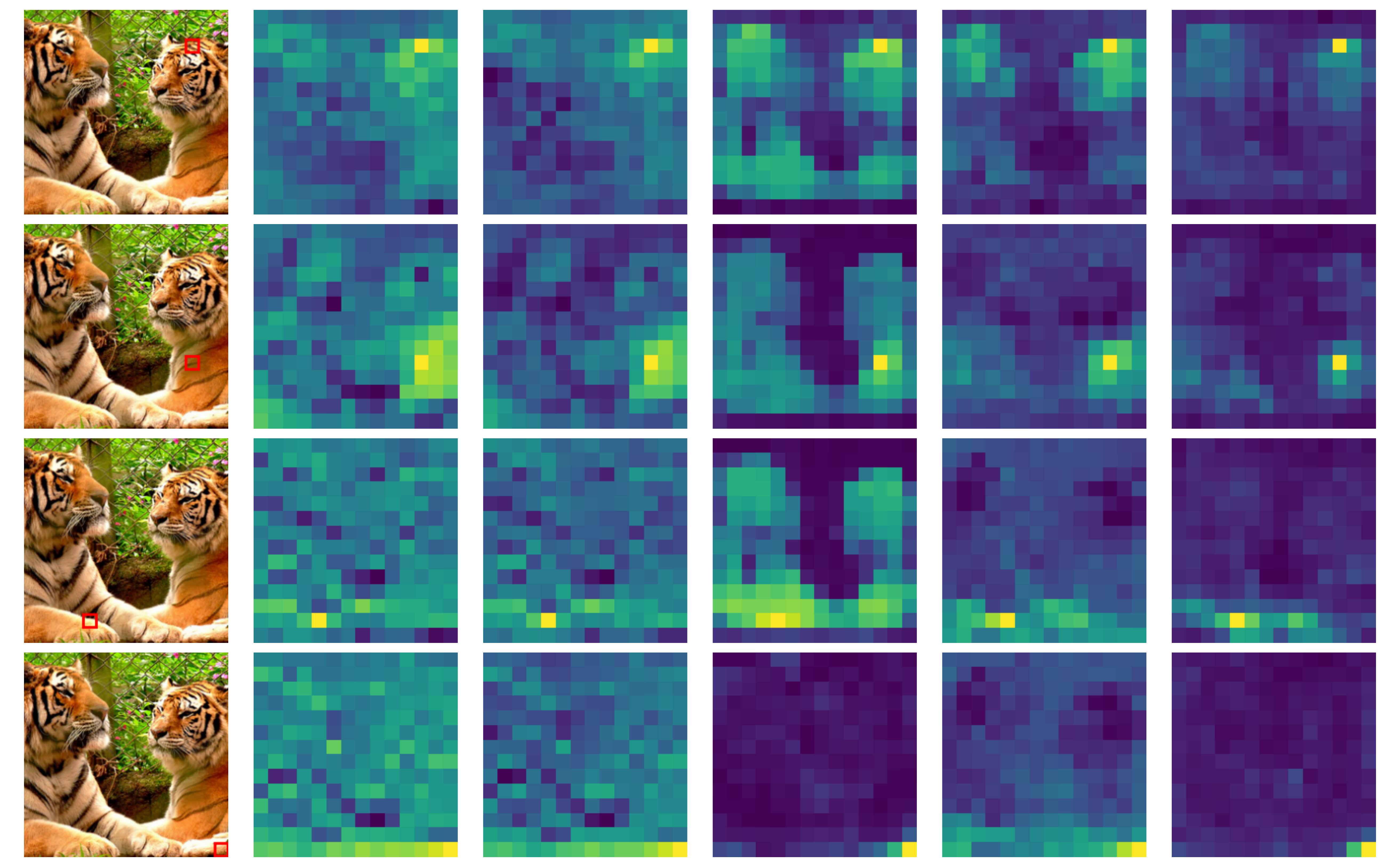}
    \caption{\textbf{Visualization of targets similarity from different teachers.} The area with a red box in the first column is the \textit{query} patch. The right 5 columns represent similarity heat-maps of different teachers. Each grid indicates its similarity with the \textit{query}.}
    \label{fig:appendix_show2}
\end{figure*}

We supply more visualization of targets similarity from different teachers as shown in \cref{fig:appendix_show1} and \cref{fig:appendix_show2}. Images are randomly selected from ImageNet-1K training set and $4$ random queries are picked for each image.

\end{document}